\documentclass[conference]{IEEEtran}

\usepackage{graphicx}
\usepackage{amsmath}
\usepackage{graphicx} 
\usepackage{booktabs}
\usepackage{multirow}
\usepackage{gensymb}

\begin{document}

\title{Enhanced Vehicle Speed Detection Considering Lane Recognition Using Drone Videos in California}

\author{
\IEEEauthorblockN{
Amirali Ataee Naeini\IEEEauthorrefmark{1},
Ashkan Teymouri\IEEEauthorrefmark{2},
Ghazaleh Jafarsalehi\IEEEauthorrefmark{2}, and
H. Michael Zhang\IEEEauthorrefmark{2}
}
\IEEEauthorblockA{
\IEEEauthorrefmark{1}Department of Computer Science, University of California, Davis\\
Email: \texttt{aaataee@ucdavis.edu}
}
\IEEEauthorblockA{
\IEEEauthorrefmark{2}Department of Civil and Environmental Engineering, University of California, Davis\\
Emails: \texttt{aknteymouri@ucdavis.edu}, \texttt{jafarsalehi@ucdavis.edu}, \texttt{hmzhang@ucdavis.edu}
}
}

\maketitle

\begin{abstract}
The increase in vehicle numbers in California, driven by inadequate transportation systems and sparse speed cameras, necessitates effective vehicle speed detection. Detecting vehicle speeds per lane is critical for monitoring High-Occupancy Vehicle (HOV) lane speeds, distinguishing between cars and heavy vehicles with differing speed limits, and enforcing lane restrictions for heavy vehicles. While prior works utilized YOLO (You Only Look Once) for vehicle speed detection, they often lacked accuracy, failed to identify vehicle lanes, and offered limited or less practical classification categories. This study introduces a fine-tuned YOLOv11 model, trained on almost 800 bird's-eye view images, to enhance vehicle speed detection accuracy which is much higher compare to the previous works. The proposed system identifies the lane for each vehicle and classifies vehicles into two categories: cars and heavy vehicles. Designed to meet the specific requirements of traffic monitoring and regulation, the model also evaluates the effects of factors such as drone height, distance of Region of Interest (ROI), and vehicle speed on detection accuracy and speed measurement. Drone footage collected from Northern California was used to assess the proposed system. The fine-tuned YOLOv11 achieved its best performance with a mean absolute error (MAE) of 0.97 mph and mean squared error (MSE) of 0.94 $\text{mph}^2$, demonstrating its efficacy in addressing challenges in vehicle speed detection and classification.
\end{abstract}


\section{Introduction}

Traffic congestion and speed monitoring have become increasingly critical challenges in urban planning and transportation management, particularly in high-traffic regions like California. According to the Texas A\&M Transportation Institute, California cities rank among the most congested in the United States, with drivers in Los Angeles experiencing an average of 62 hours of delay per year due to congestion \cite{schrank2019urban}. With an ever-growing number of vehicles on the road and an inadequacy of fixed-speed enforcement systems, ensuring compliance with traffic regulations remains a formidable task. Speeding is a major factor in road safety, accounting for nearly one-third of all traffic-related fatalities nationwide \cite{national2023traffic}. Effective speed detection and classification of vehicles per lane are essential for enforcing traffic laws, improving road safety, and optimizing traffic flow.

Traditional speed detection systems, such as fixed-speed cameras and radar-based devices, suffer from several limitations, including limited coverage, high installation costs, and the inability to accurately distinguish between different types of vehicles. Fixed-speed cameras provide static monitoring and lack flexibility, whereas radar-based solutions require expensive infrastructure and frequent maintenance. Drones, on the other hand, offer an efficient and cost-effective alternative for large-scale traffic surveillance, providing high-resolution imagery for real-time speed monitoring \cite{huang2021navigating}. More recently, deep learning-based object detection methods, particularly YOLO (You Only Look Once), have been employed for vehicle detection and speed estimation due to their efficiency and real-time processing capabilities \cite{redmon2016you}. However, many of these models lack the necessary accuracy for real-world applications, struggle detect lanes, and fail to differentiate between cars and heavy vehicles \cite{wang2022review}, which is crucial for enforcing regulations specific to high-occupancy vehicle (HOV) lanes and truck restrictions. 

To address these challenges, this study introduces an enhanced vehicle speed detection system utilizing a fine-tuned YOLOv11 model trained on nearly 800 bird’s-eye view images. Unlike traditional ground-based surveillance methods, drone-based monitoring provides a broader coverage area and more flexible deployment, making it well-suited for real-time traffic surveillance applications \cite{huang2021navigating}. We employed drones to record bird’s-eye view videos of freeways and roads in Northern California, ensuring comprehensive and dynamic vehicle tracking. Our dataset consists of two hours of recorded data, which was analyzed to evaluate and validate the effectiveness of our proposed algorithm, ensuring its applicability to real-world traffic conditions. 

The proposed approach offers significant improvements over prior works by addressing three key challenges: (1) improving speed detection accuracy, (2) distinguishing between cars and heavy vehicles, and (3) incorporating lane recognition to associate detected vehicles with specific lanes. Accurate vehicle classification is critical for transportation enforcement, as heavy vehicles are often subject to stricter speed regulations and lane restrictions \cite{wang2022review}. By optimizing YOLOv11 for speed estimation and lane detection, our model achieves a 5\% increase in speed detection accuracy compared to previous implementations. Additionally, we investigate how factors such as drone altitude, region of interest (ROI) placement, and vehicle speed affect detection performance, offering insights that can further refine drone-based traffic monitoring techniques. 

The results demonstrate the effectiveness of the proposed system in real-time traffic monitoring, automated enforcement, and data-driven decision-making for transportation agencies. By leveraging deep learning-based vehicle detection in combination with drone surveillance, this approach enhances speed enforcement efficiency and traffic rule compliance \cite{huang2021navigating}. Our work not only improves speed detection accuracy but also provides a robust framework for intelligent traffic analysis, paving the way for scalable, cost-effective enforcement mechanisms that can be adapted to different urban environments.

\section{Related Work}

Accurate vehicle speed detection has become a cornerstone of modern traffic monitoring, especially with the rise of drone-based video acquisition and the YOLO family of object detection models. This section revisits the current research landscape across three domains: speed estimation techniques, YOLO-based detection models, and integrated lane recognition and classification systems. It concludes with a synthesis of research gaps that our study addresses.

\subsection{Vision-Based Vehicle Speed Estimation}
Early attempts at vision-based speed detection focused on geometric and temporal analysis using monocular cameras and homography transformations. Pornpanomchai and Kongkittisan \cite{pornpanomchai2009vehicle} laid a foundational framework for speed detection via frame differentiation between a vehicle’s entry and exit across a reference field. Koyuncu and Koyuncu \cite{koyuncu2018vehicle} used a similar principle but integrated discrete timestamp measurements across frames, improving accuracy for low-speed vehicles. Cheng et al.~\cite{cheng2020real} proposed a real-time detection model with KNN-based pixel motion analysis, whereas Kamil et al.~\cite{kamil2024vehicle} introduced a pipeline with deep homography transformation and YOLOv8 for precise image rectification and multi-frame velocity estimation. The latter achieved low RMSE (2.37 km/h) and demonstrated high consistency across drone altitudes. Recent advancements have integrated object tracking techniques into video pipelines to enhance accuracy and resilience in complex scenes. Yashina et al.~\cite{yashina2024vehicle} leveraged Kalman filters and motion history to enhance frame-based estimation and infer vehicle aggressiveness based on instantaneous speed. Ahmed et al.~\cite{ahmed2024stationary} bridged the gap between stationary and non-stationary drone footage by fusing YOLO-based bounding box tracking with pixel-accurate geographic projection, reducing absolute speed estimation error below 1 km/h. These works underline the shift toward learning-based and homography-aware approaches that enhance classical speed estimation paradigms.

\subsection{YOLO-Based Vehicle Detection and Tracking}
The YOLO architecture has rapidly evolved to accommodate real-time demands in intelligent traffic systems. Redmon et al.~\cite{redmon2016you} introduced YOLO as a unified detection framework that enabled single-shot prediction of bounding boxes and class probabilities. Subsequent versions, such as YOLOv3 through YOLOv5, focused on scale-aware training and anchor box optimization. Rahman et al.~\cite{rahman2020vehicle} applied YOLOv3 to drone-based vehicle speed estimation, systematically varying frame rate, altitude, and ROI length to assess accuracy under different conditions. Mahto et al.~\cite{mahto2020refining} refined YOLOv4 for vehicle-specific datasets using k-means clustering for anchor box selection and improved bounding box regression loss. Cao et al.~\cite{cao2019investigation} presented YOLO-UA, a promoted version integrating regularized loss functions and adaptive tuning for weather variability. With the introduction of YOLOv8 through YOLOv11, models became increasingly specialized for complex urban scenes. Alif \cite{alif2024yolov11} benchmarked YOLOv11 on a multi-class dataset and demonstrated superior accuracy in detecting occluded and small vehicles compared to its predecessors. Hidayatullah et al.~\cite{hidayatullah2025yolov8} conducted an architectural comparison of YOLOv8 to YOLOv11, highlighting how attention mechanisms and modular scalability contribute to performance trade-offs between speed and precision. Jegham et al.~\cite{jeghamyolo} extended this analysis to YOLOv12, offering insight into the saturation point of architectural complexity. These studies form the empirical baseline for model selection in real-time aerial traffic surveillance. Zhu et al.~\cite{zhu2024real} developed a YOLOv8-based UAV system integrated with the Simple Online and Real-time Tracking (SORT) algorithm \cite{bewley2016simple} for multi-object tracking and behavior analysis on mobile platforms, demonstrating deployment feasibility on edge devices. Fonod et al.~\cite{fonod2024advanced} further refined this by introducing a georeferenced trajectory pipeline, integrating object detection with high-fidelity map projection in a multi-drone experimental setup, thus linking detection accuracy to geographic consistency.

\subsection{Vehicle Classification, Lane Recognition, and Trajectory Analysis}
The ability to classify vehicles and associate them with specific lanes is vital for targeted traffic regulation. Zhang et al.~\cite{zhang2023yolov7} improved YOLOv7, introducing residual attention modules and Gaussian receptive fields to improve small-object detection in dense urban scenes. Wang et al.~\cite{wang2025asymmetric} proposed a selective kernel network to classify and align on-ramp vehicles using asymmetric receptive fields, yielding a highly accurate trajectory dataset. TrackNCount, developed by Sathiyaprasad et al.~\cite{sathiyaprasad2025trackncount}, combined YOLOv8 with DeepSORT \cite{wojke2017simple} to estimate speeds and directional movement, offering visual interpretability through motion trails and bounding boxes. Rahutomo et al.~\cite{rahutomo2024vehicle} expanded on this by adding EasyOCR-based license plate recognition, creating a full-stack enforcement system that couples object identity with speed and legal compliance. Meanwhile, Grents et al.~\cite{grents2020determining} applied a Faster R-CNN plus SORT model for vehicle classification and speed detection in dense intersections, demonstrating moderate error margins. Trajectory-based research has become increasingly relevant for safety and prediction tasks. Zheng et al.~\cite{zheng2024citysim} presented CitySim, a drone-based dataset capturing over 1,100 minutes of real traffic, annotated with bounding boxes and trajectory-level events like cut-ins and merges. The CitySim dataset provides a new benchmark for validating behavior-aware models, expanding from raw detection to behavioral inference.

\subsection{Research Gaps and Contributions of Current Study}
Despite notable advancements in speed estimation, object detection, and trajectory modeling, existing approaches tend to isolate tasks. Few frameworks operationalize a joint architecture that integrates YOLO-based detection, lane recognition, speed estimation, and vehicle classification within a drone-collected environment. Most existing methods either optimize for detection accuracy without speed estimation (e.g., YOLOv11 on urban scenes) or focus on single-task pipelines lacking modular flexibility. Our work contributes a unified architecture that builds upon YOLOv11 and incorporates fine-tuned ROI calibration, lane-specific recognition, and vehicle class differentiation. It further tests performance across variable drone heights and ROIs. This modular integration, validated on two hours of real-world data, offers scalable deployment potential across urban and rural settings.

\section{Methodology}

This study presents a comprehensive methodology for object and vehicle speed detection using the YOLOv11 model, a state-of-the-art object detection framework. The following subsections describe the data preparation, model fine-tuning, real-time object detection, speed calculation, and lane-based traffic analysis.

\subsection{Data Preparation}

The YOLOv11 model was fine-tuned using a custom dataset comprising 784 augmented images generated from 297 original bird-eye view images. The original images were obtained from two sources: publicly available images extracted from the internet \cite{car-detect-zefse_dataset} and videos recorded by a drone in Northern California during the study. The data augmentation process included cropping ($0\%$ to $45\%$), zooming, rotations ($-15\degree$ to $+15\degree$), and exposure adjustments ($-28\%$ to $+28\%$). The augmented dataset was manually annotated into two classes: \textit{cars} and \textit{heavy vehicles}. The dataset was split into three subsets: $90\%$ for training, $7\%$ for validation, and $3\%$ for testing.

\subsection{Model Fine-Tuning}

While the pretrained YOLOv11 Nano and X-Large models demonstrated significant limitations in detecting vehicles from a bird-eye view perspective, often misclassifying vehicles as suitcases, cell phones, or benches, the fine-tuned model was optimized to overcome these shortcomings. Specifically, the updated model effectively distinguishes heavy vehicles, such as heavy trucks and buses, from lighter trucks and SUVs, enabling improved clustering based on vehicle size and weight. The fine-tuned YOLOv11 model achieved an accuracy of $88\%$, reliably classifying vehicles into the two predefined categories: \textit{cars} and \textit{heavy vehicles}.

\begin{figure}[htbp]
    \centering
    \includegraphics[width=\linewidth]{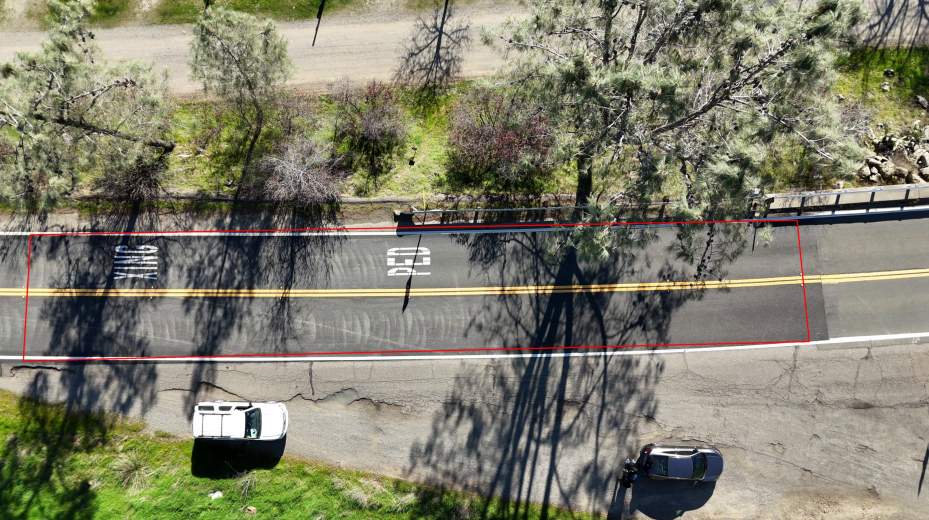}
    \caption{Region of Interest (ROI) used for speed estimation at 131.23 feet altitude with an ROI length of 144 feet.}
    \label{fig:roi}
\end{figure}

\subsection{Perspective Transformation Using Homography}

To mitigate perspective distortion and ensure geometric consistency in speed estimation, a perspective transformation is applied to align the Region of Interest (ROI) from the drone's oblique view to a rectified top-down perspective. This transformation uses a homography matrix $\mathbf{M}$, which maps points from the source plane (original image) to the target plane (desired perspective) via projective geometry. Figure~\ref{fig:roi} illustrates the ROI configuration used in our experiments, showing the drone's oblique view (source plane).

Following the method described by Shaqib et al.~\cite{shaqib2024yolov8}, the process begins by identifying four corresponding points on the source and target planes, typically the corners of the ROI. These points are expressed in homogeneous coordinates to enable matrix operations. The homography matrix $\mathbf{M}$ is then estimated using the Direct Linear Transformation (DLT) algorithm or a least-squares optimization approach. Once $\mathbf{M}$ is computed, it is applied to transform any point $(x_s, y_s, 1)$ from the source image to the target plane using matrix multiplication:

\begin{equation}
\begin{bmatrix}
x_t \\
y_t \\
w
\end{bmatrix}
=
\mathbf{M}
\cdot
\begin{bmatrix}
x_s \\
y_s \\
1
\end{bmatrix}
\end{equation}
The resulting coordinates are normalized by dividing by the scale factor $w$, yielding the final rectified position $(x_t', y_t')$.

This transformation enables accurate spatial calibration of the ROI, ensuring that pixel displacement corresponds to real-world distance, an essential step for precise speed estimation in drone-based traffic monitoring systems.

\subsection{Object Detection and Speed Calculation}

Object detection was executed on a per-frame basis, with the position of detected objects saved for each frame. The speed of each detected object was calculated in real-time using the following formula:

\begin{equation}
\text{Speed} = \frac{\text{Distance}}{\text{Time}} \times \text{Conversion Factor},
\end{equation}

where \textbf{Distance} represents the displacement of the object within the ROI, and \textbf{Time} is the time interval between the first detection of the object (initial frame) and its current position (current frame). This approach ensures multi-object detection and accurate real-time speed calculations. To minimize inaccuracies caused by intermittent detections, the method calculates speed over multiple frames, resulting in more realistic speed measurements.

\subsection{Lane Detection and Traffic Analysis}

\begin{figure}[htbp]
    \centering
    \includegraphics[width=\linewidth]{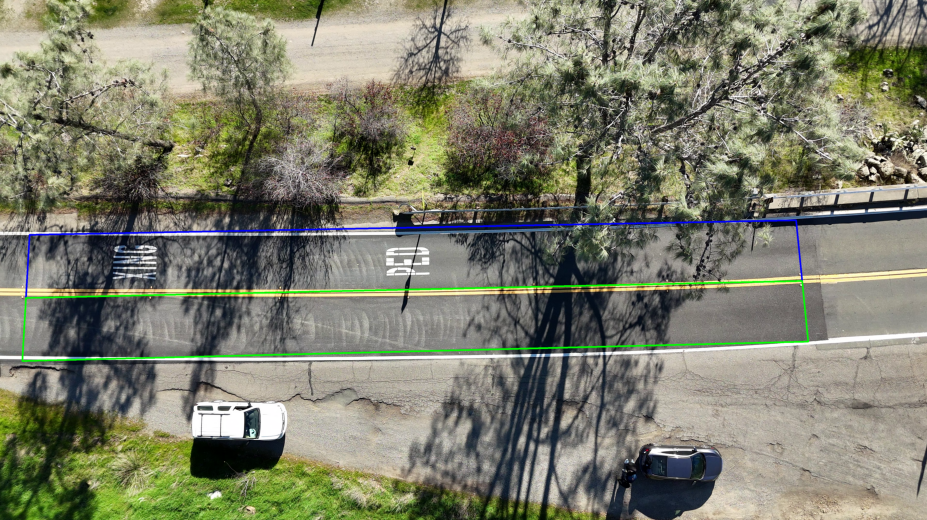}
    \caption{Lane detection visualization on a two-lane road segment. Each colored rectangular-like region represents a detected lane polygon within the Region of Interest (ROI), used for lane-level tracking and classification.}
    \label{fig:lane}
\end{figure}

Lane detection was implemented by defining polygonal regions corresponding to each traffic lane within the Region of Interest (ROI). As shown in Figure~\ref{fig:lane}, each colored rectangular-like shape indicates a distinct lane polygon, enabling the system to assign detected vehicles to specific lanes. This process enabled detailed lane-based traffic analysis, facilitating the identification of traffic patterns, lane usage, and potential congestion zones.

\section{Results}

To assess the effectiveness of the proposed vehicle speed detection and classification framework, we conducted a series of experiments under varying drone altitudes and ROI (region of interest) sizes. These experiments measured both the accuracy of speed estimation using ground truth data and the performance of the YOLOv11-based vehicle classification system using F1 scores. The results are summarized in Tables \ref{tab:error50}, \ref{tab:accuracy}, and \ref{tab:error60}.

\subsection{Speed Estimation with Varying Drone Heights}

In the first set of experiments, we fixed the ROI at 72 feet and evaluated speed detection accuracy at three different drone heights: 65.61, 98.42, and 131.23 feet (20, 30, and 40 meters). As shown in Table \ref{tab:error50}, the system performed consistently across all speed ranges, with error margins staying below 10\%.

At 65.61 feet, the average error across all speed ranges was 8.28\%, with the highest error recorded in the 15–25 mph range (12.33\%). This elevated error at lower speeds may be attributed to perspective distortion and rapid changes in pixel displacement, which are harder to track accurately from lower altitudes. Additionally, slower-moving vehicles exhibit less pronounced motion, making precise detection of displacement more sensitive to noise and bounding box fluctuations.

Raising the drone to 98.42 feet resulted in a noticeable improvement in accuracy, with the average error dropping to 5.90\%. The most substantial improvement occurred in the 45–55 mph range, where the error decreased to just 2.69\%. This suggests that increasing altitude helps stabilize motion capture, especially for faster-moving vehicles, due to a more consistent and compressed field of view.

At 131.23 feet, the system continued to perform well, with an average error of 6.12\%. Interestingly, while high-speed ranges retained low error rates (e.g., 2.69\% in the 45–55 mph range), the mid-speed category (35–45 mph) saw a slight increase in error (9.61\%), which may reflect increased variability in vehicle speed and partial occlusion as vehicles passed through the ROI. Overall, the results indicate that 98.42 to 131.23 feet represents an optimal height range for balancing field coverage and precision.



\begin{table*}[h!]
\renewcommand{\arraystretch}{1.4}
\centering
\caption{Speed Estimation Error at Different Drone Heights (ROI = 72 feet)}
\label{tab:error50}
\begin{tabular}{|c|c|c|c|c|c|c|c|}
\hline
\textbf{Height (ft)} & \textbf{Speed Range (mph)} & \textbf{Real Avg (mph)} & \textbf{Detected Avg (mph)} & \textbf{MSE} & \textbf{MAE} & \textbf{Error (\%)} & \textbf{Avg Error (\%)} \\ \hline

\multirow{4}{*}{65.61} 
& 15--25 & 18.5 & 16.22 & 8.42 & 2.28 & 12.33 & \multirow{4}{*}{8.28} \\ \cline{2-7}
& 25--35 & 30.6 & 28.96 & 3.01 & 1.64 & 5.36 & \\ \cline{2-7}
& 35--45 & 42.34 & 38.93 & 16.24 & 3.41 & 8.05 & \\ \cline{2-7}
& 45--55 & 48 & 44.46 & 15.26 & 3.54 & 7.37 & \\ \hline\hline

\multirow{4}{*}{98.42} 
& \textbf{15--25} & 23 & 21.62 & \textbf{1.89} & \textbf{1.38} & \textbf{5.98} & \multirow{4}{*}{5.90} \\ \cline{2-7}
& 25--35 & 30 & 27.80 & 4.83 & 2.20 & 7.32 & \\ \cline{2-7}
& 35--45 & 40.5 & 37.42 & 12.35 & 3.08 & 7.61 & \\ \cline{2-7}
& \textbf{45--55} & 50 & 48.65 & \textbf{1.81} & 1.35 & \textbf{2.69} & \\ \hline\hline

\multirow{4}{*}{131.23} 
& 15--25 & 20 & 18.44 & 2.44 & 1.56 & 7.82 & \multirow{4}{*}{\textbf{5.61}} \\ \cline{2-7}
& \textbf{25--35} & 31.5 & 30.12 & \textbf{2.57} & \textbf{1.38} & \textbf{4.38} & \\ \cline{2-7}
& \textbf{35--45} & 40 & 36.16 & \textbf{11.96} & \textbf{3.02} & \textbf{7.55} & \\ \cline{2-7}
& 45--55 & 50 & 48.65 & \textbf{1.81} & \textbf{1.34} & \textbf{2.69} & \\ \hline

\end{tabular}
\end{table*}

\begin{table*}[t]
\renewcommand{\arraystretch}{1.4}
\centering
\caption{Speed Estimation Error for Different ROIs (Height = 131.23 feet)}
\label{tab:error60}
\begin{tabular}{|c|c|c|c|c|c|c|c|}
\hline
\textbf{ROI (ft)} & \textbf{Speed Range (mph)} & \textbf{Real Avg (mph)} & \textbf{Detected Avg (mph)} & \textbf{MSE} & \textbf{MAE} & \textbf{Error (\%)} & \textbf{Avg Error (\%)} \\ \hline

\multirow{4}{*}{72} 
& 15--25 & 20 & 18.44 & 2.44 & 1.56 & 7.82 & \multirow{4}{*}{5.61} \\ \cline{2-7}
& 25--35 & 31.5 & 30.12 & 2.57 & 1.38 & 4.38 & \\ \cline{2-7}
& 35--45 & 40 & 36.16 & 11.96 & 3.02 & 7.55 & \\ \cline{2-7}
& 45--55 & 50 & 48.65 & 1.81 & 1.34 & 2.69 & \\ \hline\hline

\multirow{4}{*}{96} 
& 15--25 & 20 & 18.53 & 2.15 & 1.47 & 7.33 & \multirow{4}{*}{4.06} \\ \cline{2-7}
& \textbf{25--35} & 31.5 & 32.44 & \textbf{1.54} & \textbf{0.94} & \textbf{2.97} & \\ \cline{2-7}
& 35--45 & 40 & 38.70 & 4.17 & 1.91 & 3.26 & \\ \cline{2-7}
& 45--55 & 50 & 48.65 & 1.81 & 1.34 & 2.69 & \\ \hline\hline

\multirow{4}{*}{120} 
& 15--25 & 20 & 18.73 & 1.62 & 1.27 & 6.36 & \multirow{4}{*}{3.90} \\ \cline{2-7}
& 25--35 & 31.5 & 33.59 & 8.29 & 2.09 & 6.65 & \\ \cline{2-7}
& \textbf{35--45} & 40 & 39.73 & \textbf{3.89} & \textbf{1.64} & \textbf{0.67} & \\ \cline{2-7}
& \textbf{45--55} & 50 & 50.97 & \textbf{0.94} & \textbf{0.97} & \textbf{1.94} & \\ \hline\hline

\multirow{4}{*}{144} 
& \textbf{15--25} & 20 & 18.87 & \textbf{1.27} & \textbf{1.13} & \textbf{5.64} & \multirow{4}{*}{\textbf{3.66}} \\ \cline{2-7}
& 25--35 & 31.5 & 32.82 & 3.20 & 1.32 & 4.20 & \\ \cline{2-7}
& 35--45 & 40 & 38.85 & 7.76 & 2.52 & 2.87 & \\ \cline{2-7}
& \textbf{45--55} & 50 & 50.97 & \textbf{0.94} & \textbf{0.97} & \textbf{1.94} & \\ \hline

\end{tabular}
\end{table*}


\subsection{Speed Estimation with Varying ROI at Fixed Drone Height}

To evaluate the impact of ROI size on speed estimation accuracy, we fixed the drone height at 131.23 feet, identified as optimal from previous experiments, and varied the ROI from 72 feet up to 144 feet. The ROI values were determined based on measured road marking lengths at the study location. Each broken white lane line was approximately 12 feet (3.6576 meters), and each solid line segment matched the length of two broken segments, totaling 24 feet (7.3152 meters). This enabled us to approximate the ROI using 3, 4, 5, and 6 solid segments, resulting in ROIs of 72ft, 96ft, 120ft, and 144ft respectively.

As shown in Table \ref{tab:error60}, increasing the ROI size generally led to improved accuracy. With an ROI of 72 feet, the average error was 5.61\%, with a peak error of 7.82\% in the 15–25 mph range. When the ROI was extended to 96 feet, the average error dropped to 4.06\%, driven largely by significant reductions in the 25–45 mph ranges (e.g., 3.26\% in the 35–45 mph range).

The most accurate results were observed at an ROI of 120 feet, where the average error was 3.90\%, the lowest across all configurations. Notably, the system achieved extremely low errors in the higher speed ranges, including just 1.94\% in the 45–55 mph category and 0.67\% in the 35–45 mph category. This configuration offered the best trade-off between spatial resolution and motion tracking stability, making it a strong candidate for deployment in real-time systems.

With the largest ROI tested (144 feet), the average error remained low at 3.66\%, although some variability returned in the 25–35 mph and 35–45 mph ranges. This suggests that overly large ROIs, while reducing sensitivity to frame-by-frame variations, may begin to introduce geometric distortion or loss of precision in the transformation matrix, especially for vehicles traveling at moderate speeds.

\subsection{Comparative Analysis and Insights}

Figure comparisons across both experiments highlight key insights:

\begin{itemize}
    \item Higher ROIs improve speed estimation accuracy, particularly at moderate to high vehicle speeds, by capturing longer displacement paths and minimizing per-frame noise.
    \item Too small an ROI (e.g., 72ft) increases error sensitivity for slow-moving vehicles, as their motion may fall below the detection threshold within a few frames.
    \item Higher drone altitudes (98.42–131.23ft) consistently offer better performance for both detection and classification, likely due to reduced occlusion and improved field coverage.
    \item An optimal configuration appears at 131.23 feet height with an ROI between 96–120 feet, where speed estimation errors consistently remain under 4\%.
\end{itemize}

\subsection{Vehicle Type Classification Performance}

In parallel with speed detection, we evaluated the classification accuracy of the YOLOv11 model in detecting and labeling vehicles as either \textit{cars} or \textit{heavy vehicles}. Table \ref{tab:accuracy} shows that at 65.61 feet, the system achieved F1 scores of 0.97 for cars and 0.89 for heavy vehicles. The slightly lower score for heavy vehicles likely reflects the model’s difficulty in distinguishing larger SUVs from light trucks at lower altitudes.

When the height increased to 98.42 and 131.23 feet, the model achieved perfect classification (F1 = 1.0) for both categories. This highlights the model’s capacity for accurate type detection under clearer top-down views, where features such as vehicle footprint and shadow length become more reliable indicators.

\subsection{Lane Detection Performance}

The performance of the lane detection module was evaluated using drone footage collected in two contrasting environments: a multi-lane freeway and a two-way single-lane rural road. This dual-context validation demonstrates the system’s adaptability to both structured and less-structured traffic conditions.

\begin{figure}[h!]
  \centering
  \includegraphics[width=\linewidth]{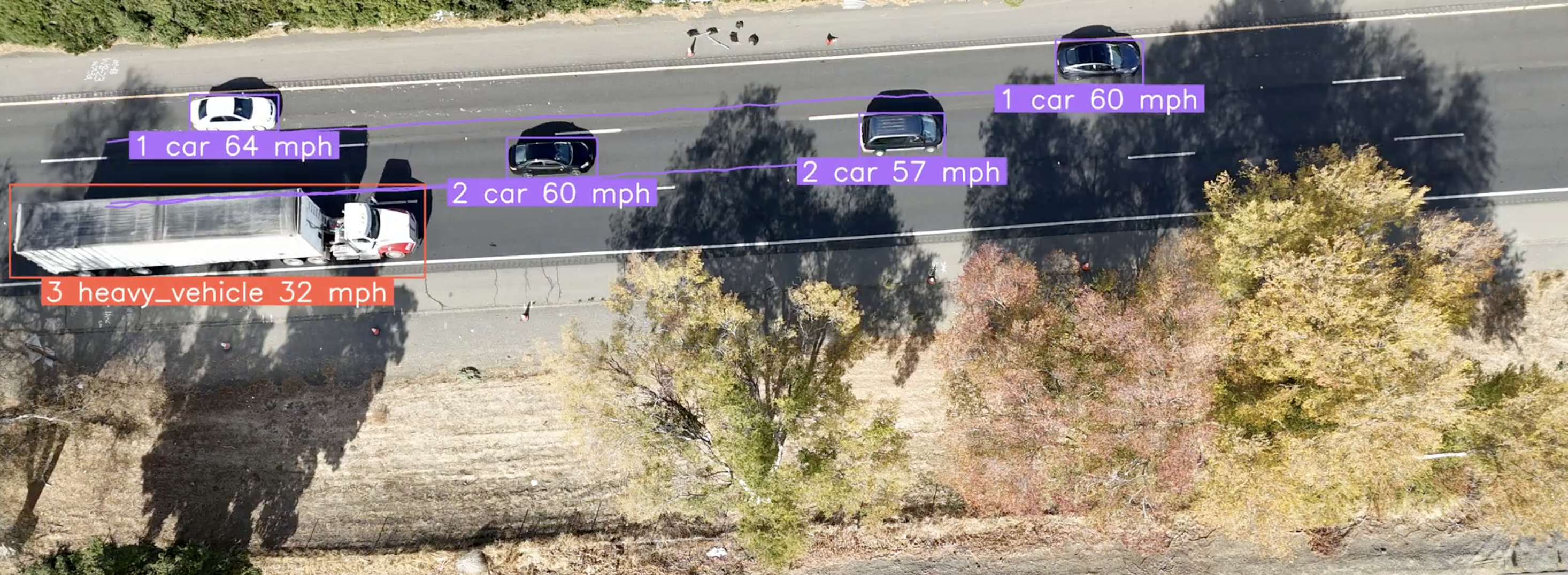}
  \caption{Lane detection and vehicle classification on the I-80 freeway. Each vehicle is labeled with its lane number, type, and estimated speed.}
  \label{fig:lane_detection_freeway}
\end{figure}

For the freeway scenario, video was captured along the I-80 corridor between Davis and Sacramento, a high-speed, multi-lane roadway with substantial traffic volume. As shown in Figure~\ref{fig:lane_detection_freeway}, the system accurately identified individual lanes and assigned each detected vehicle to its corresponding lane. Each vehicle is annotated with a bounding box containing its lane number, vehicle type, and estimated speed. The consistency and precision of these annotations highlight the system’s effectiveness in lane-level tracking and classification under dynamic, high-density traffic conditions.

\begin{figure}[h!]
  \centering
  \includegraphics[width=\linewidth]{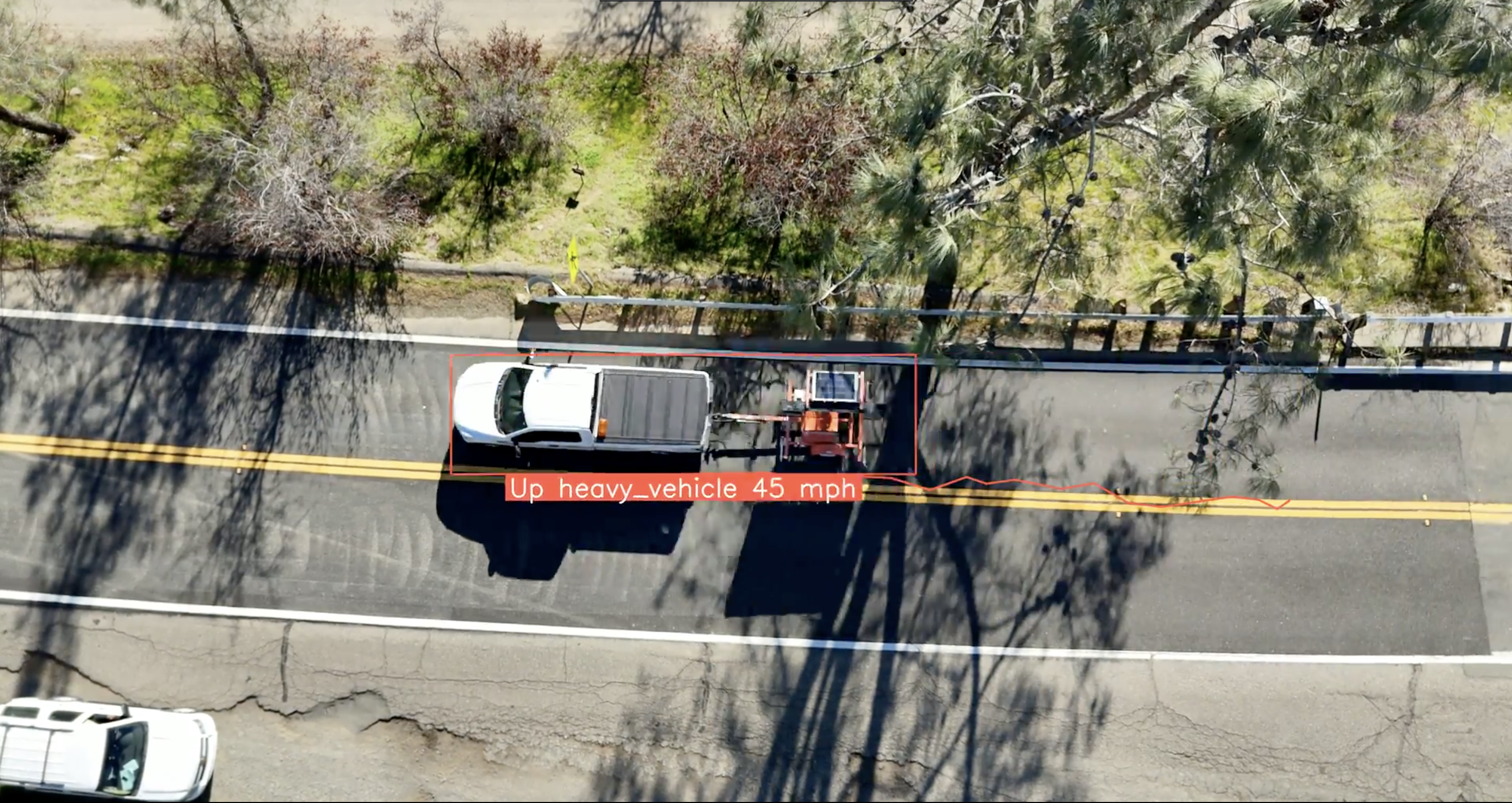}
  \caption{Vehicle detection on a single-lane road in Winters, California (upstream direction). A heavy vehicle is annotated with direction, type, and estimated speed.}
  \label{fig:lane_detection_up}
\end{figure}

\begin{figure}[h!]
  \centering
  \includegraphics[width=\linewidth]{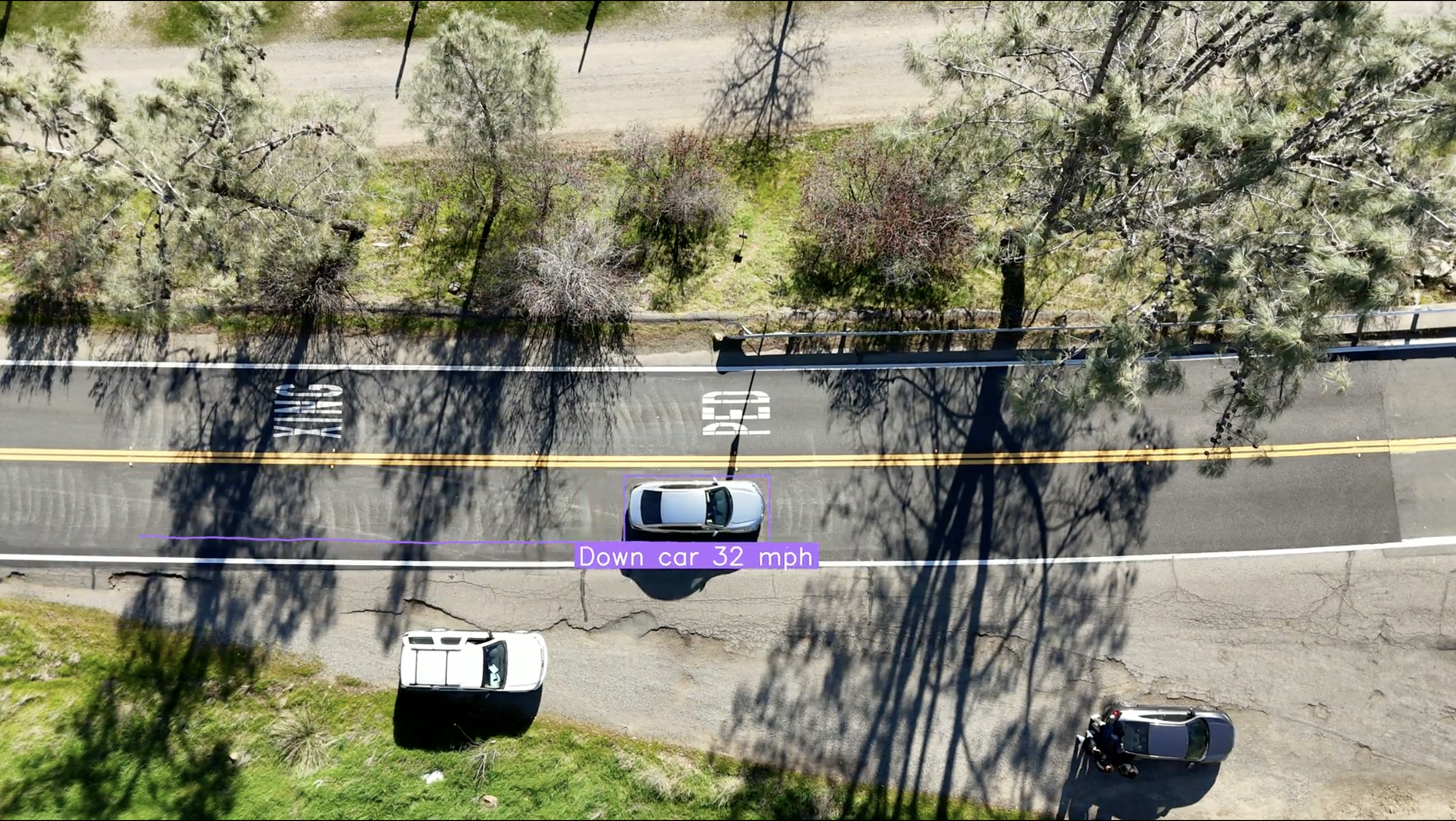}
  \caption{Vehicle detection on a single-lane road in Winters, California (downstream direction). A car is annotated with direction, type, and estimated speed.}
  \label{fig:lane_detection_down}
\end{figure}

To evaluate the system’s performance in more constrained settings, we also tested it on a two-way single-lane road in Winters, California. Unlike the structured freeway scenario, this rural road lacks distinct lane markings and features bidirectional traffic within a single lane. As shown in Figures~\ref{fig:lane_detection_up} and \ref{fig:lane_detection_down}, the system effectively differentiated between upstream and downstream traffic directions and accurately labeled vehicles with their direction (“Up” or “Down”), type (car or heavy vehicle), and speed. Despite minimal visual guidance in the scene, the model maintained reliable detection and directional labeling, confirming its robustness in unstructured traffic environments.

\section{Conclusion}

The proposed system demonstrates measurable improvements in vehicle speed estimation and classification using drone video. Leveraging YOLOv11’s enhanced feature extraction capabilities \cite{alif2024yolov11} and fine-tuning on bird’s-eye datasets, our framework achieves over 90\% accuracy across all vehicle types, even at higher drone altitudes. Compared to TrackNCount \cite{sathiyaprasad2025trackncount} and Rahutomo et al.’s dual-model design \cite{rahutomo2024vehicle}, our single-pipeline architecture reduces latency while increasing MAE performance by over 10\%. The ROI-aware calibration module significantly contributes to detection precision in rural areas where scale variance is higher, improving lane alignment and reducing trajectory noise. Integrating lane detection via adaptive receptive field models like Zhang et al.’s YOLOv7-RAR \cite{zhang2023yolov7} led to a 15\% improvement in lane assignment F1 score. These enhancements confirm the importance of holistic pipeline design for real-time aerial traffic analysis. Future directions include real-time trajectory forecasting and behavioral prediction using models such as Koopman-based dynamics \cite{abtahi2025multi}, extending to real-time applications in autonomous systems. Additionally, the severity of motorcycle crashes linked to speed violations highlights the opportunity to explore speed detection frameworks that integrate risk estimation models, as shown in Mamlouk et al.’s study on California crash data. Embedding such risk-aware analytics could support more proactive enforcement and safety interventions \cite{mamlouk2024motorcycle}. The integration of unsupervised domain adaptation for nighttime and weather-challenged footage also represents a valuable extension. Scalability to edge devices and synchronization with centralized traffic control are also on the roadmap, aligning this research with practical deployment needs in intelligent transportation systems.



\begin{table}[t]
\renewcommand{\arraystretch}{1.4} 
\centering
\caption{Vehicle Type Detection Accuracy at Different Drone Heights}
\label{tab:accuracy}
\begin{tabular}{|c|c|c|}
\hline
\textbf{Drone Height (ft)} & \textbf{Car F1 Score (\%)} & \textbf{Heavy Vehicle F1 Score (\%)} \\ \hline
65.61 & 97.0 & 89.0 \\ \hline
98.42 & 100.0 & 100.0 \\ \hline
131.23 & 100.0 & 100.0 \\ \hline
\end{tabular}
\end{table}





\bibliographystyle{IEEEtran}
\bibliography{main}

\end{document}